\documentclass[journal]{IEEEtran}
\ifCLASSINFOpdf
\else
\fi
\usepackage{graphicx}
\usepackage[font=footnotesize,labelfont=bf,skip=0pt]{caption}
\usepackage{epstopdf}
\usepackage{amssymb}
\usepackage{amsmath} 
\usepackage{enumerate}
\usepackage{etaremune}
\usepackage{setspace}
\usepackage{comment}
\usepackage{color}
\usepackage[table]{xcolor}
\usepackage{soul,xcolor}
\setstcolor{red}
\usepackage{wrapfig}
\usepackage[algo2e]{algorithm2e} 
\usepackage{algorithm} 
\usepackage{algpseudocode} 
\usepackage[hidelinks]{hyperref}
\usepackage{cite}
\definecolor{blue0}{RGB}{100, 211, 100}
\definecolor{red}{RGB}{150, 11, 23}
\usepackage{soul}
\usepackage{booktabs}
\usepackage{float}
\usepackage{amsmath}
\usepackage{subfigure}
\usepackage{pgfplots} 
\pgfplotsset{compat=newest} 
\pgfplotsset{plot coordinates/math parser=false} 
\newlength\figureheight 
\newlength\figurewidth 
\usetikzlibrary{mindmap}
\usepackage{placeins}
\usetikzlibrary{arrows}
\usepgfplotslibrary{patchplots}

\definecolor{UBblue}{RGB}{0, 111, 213}
\definecolor{UBred}{RGB}{150, 11, 23}

 \usepackage{soul}
 \usepackage{tcolorbox}
\newtcolorbox{mybox}{colback=yellow!50!white,colframe=yellow!100!black}
 \usepackage[switch]{lineno}

\def\subparagraph{}
\usepackage{titlesec}
\titlespacing{\subsection}{0pt}{*1}{*1}

\renewcommand{\thesubsubsection}{\arabic{subsubsection}}

\titleformat{\subsubsection}[runin]{\itshape}{\thesubsubsection)}{1em}{}
\titlespacing*{\subsubsection}{\parindent}{0pt}{*1}
 
\begin{document}

\title{\LARGE{{\bf Integrating Uncertainty-Aware Human Motion Prediction into Graph-Based Manipulator Motion Planning}}}

\author{Wansong Liu$^{1}$, Kareem Eltouny$^{2}$, Sibo Tian$^{3}$, Xiao Liang$^{4}$, Minghui Zheng$^{3}$
    \thanks{This work was supported by the USA National Science Foundation  (Grants: 2026533/2422826 and 2132923/2422640). This work involved human subjects or animals in its research. The authors confirm that all human/animal subject research procedures and protocols are exempt from review board approval}
	\thanks{$^{1}$ Wansong Liu is with the Mechanical and Aerospace Engineering Department, University at Buffalo, Buffalo, NY 14260, USA.		{\tt\small Email: wansongl@buffalo.edu}.}
	\thanks{$^{2}$ Kareem Eltouny is with the Civil, Structural and Environmental Engineering Department, University at Buffalo, Buffalo, NY14260, USA. {\tt\small Email: keltouny@buffalo.edu}.}
   	\thanks{$^{3}$ Sibo Tian and Minghui Zheng are with the J. Mike Walker '66 Department of Mechanical Engineering, Texas A\&M University, College Station, TX 77840, USA. {\tt\small Email:  \{sibotian, mhzheng\}@tamu.edu}.}
 \thanks{$^{4}$ Xiao Liang is with the Zachry Department of Civil \& Environmental Engineering, Texas A\&M University, College Station, TX 77840, USA. {\tt\small Email: xliang@tamu.edu}.}
		\thanks{Correspondence to Minghui Zheng and Xiao Liang.}
}

\maketitle
\begin{abstract}
There has been a growing utilization of industrial robots as complementary collaborators for human workers in re-manufacturing sites. Such a human-robot collaboration (HRC) aims to assist human workers in improving the flexibility and efficiency of labor-intensive tasks. In this paper, we propose a human-aware motion planning framework for HRC to effectively compute collision-free motions for manipulators when conducting collaborative tasks with humans. We employ a neural human motion prediction model to enable proactive planning for manipulators. Particularly, rather than blindly trusting and utilizing predicted human trajectories in the manipulator planning, we quantify uncertainties of the neural prediction model to further ensure human safety. Moreover, we integrate the uncertainty-aware prediction into a graph that captures key workspace elements and illustrates their interconnections. Then a graph neural network is leveraged to operate on the constructed graph. Consequently, robot motion planning considers both the dependencies among all the elements in the workspace and the potential influence of future movements of human workers. We experimentally validate the proposed planning framework using a 6-degree-of-freedom manipulator in a shared workspace where a human is performing disassembling tasks. The results demonstrate the benefits of our approach in terms of improving the smoothness and safety of HRC. A brief video introduction of this work is available via \href{https://zh.engr.tamu.edu/wp-content/uploads/sites/310/2024/05/Integrate_prediction_into_planning.mp4}{\textcolor{gray}{\underline{link}}}.
\end{abstract}


  \vspace{-0.1in}
\section{Introduction}
To facilitate efficient and safe disassembly, robots are usually employed as complementary collaborators to work closely with human operators \cite{lee2022robot,lee2024review}. In such close collaboration, robots are required to generate collision-free motions and adjust their motions efficiently. The planning problem turns out to be complicated when human operator's behaviors are involved since real-time responsiveness necessitates quick motion generation in the constantly changing configuration space. 
Manipulators must respond adaptively to human operator's actions. Predicting human motions allows collaborative robots to proactively plan motions, ensuring a safe and seamless human-robot collaboration (HRC) \cite{liu2017human}.

Integrating human motion prediction into robotic motion planning has two technical challenges. One is that human motion is inherently complex and stochastic, which requires robust prediction models to handle the uncertainties arising from human behavior variations or unexpected actions \cite{fridovich2020confidence, tian2023transfusion, eltouny2024tgn}. Addressing such uncertainties holds particular importance in terms of ensuring the safety in HRC as unreliable predictions can potentially result in the planning of a dangerous trajectory. The second challenge is that the prediction as well as the uncertainty introduce additional computational complexity for generating robot motions \cite{unhelkar2018human}. To have a real-time responsiveness in HRC, the planning algorithms must integrate the prediction and the uncertainty in an efficient way and find collision-free motions within tight time constraints.

\begin{figure*}[h]

	\centering 
	\includegraphics[width=1.4\columnwidth]{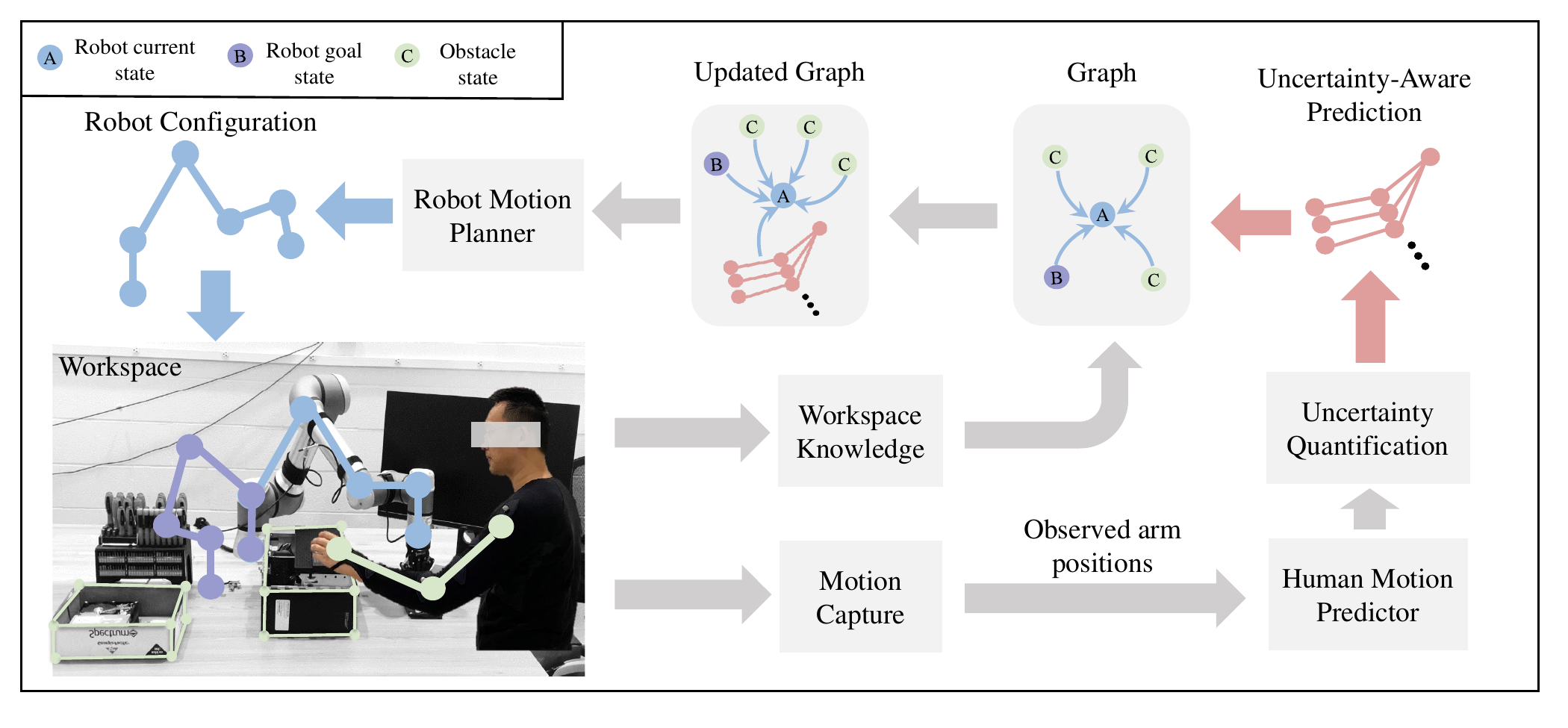}

	\caption{Overview of the proposed HRC motion planning framework: 1) Observed human motions are used to compute future uncertainty-aware human motions; 2) The collaboration workspace is converted to a graph. It preserves the structural attributes of objects by employing multiple nodes and edges. The key elements and characteristics of the workspace are depicted through nodes with features, and their connections are established through edges. Note that we only show a few of the nodes in the figure to simplify the illustration. 3) The uncertainty-aware prediction is also represented using nodes and edges and naturally integrated into the overall graph. 4) The GNN-based motion planner eventually generates a safe robot configuration, directing the robot toward the desired goal, while avoiding the moving human agent in close vicinity.
	} \label{fig:overview}
 \vspace{-0.2in}
\end{figure*}

In this paper, we propose a motion planning framework to enhance the seamless and safe collaboration between humans and manipulators in disassembly processes. Fig.~\ref{fig:overview} shows the overview of the framework. The framework is comprised of two modules. The first one is the uncertainty-aware human motion prediction. It seeks to provide future trajectories of human operators and the uncertainty of the network-based prediction model for the purpose of safe manipulator motion planning. The second module is a graph-based neural motion planner that incorporates uncertainty-aware prediction and generates collision-free manipulator motions. We transform the collaboration workspace into a graph representation that encapsulates the relationships and dependencies among the objects within the workspace. The uncertainty-aware predictions are represented as nodes and edges, which are intuitively integrated into the constructed graph and interconnected with other objects.

In summary, the main contributions of this work are summarized as follows:
\begin{itemize}
\item We present a framework for HRC that naturally integrates the motion planning of high-DOF robot manipulators and uncertainty-aware human motion prediction, using graph neural networks.
\item The inherent uncertainty of the human motion prediction model is incorporated into the robot motion planning intuitively and conveniently (i.e., using nodes and edges) to enhance safety in HRC. 
\item We conduct comprehensive experimental studies within a collaborative disassembly scenario to validate the performance of our model. The proposed planning framework showcases the benefits in terms of earlier robot's response and near-optimal trajectory planning when a sudden human intervention occurs.
\end{itemize}

\section{Related Works}
\subsection{Human motion prediction}

Traditional statistic-based models have been utilized to learn the probability distribution of human motion, enabling them to reason about possible future human trajectories based on historical data, such as the hidden Markov model \cite{moudoud2022detection} and the Gaussian regression model \cite{park2019planner}. Although these probabilistic methods are suited for capturing the stochastic nature of human motion, their performance tends to be less satisfactory when dealing with intricate motion patterns. To predict complex human motion, recurrent neural networks (RNNs) have been widely used to obtain deterministic future human trajectories \cite{liu2022dynamic}. In addition, graph convolutional networks \cite{li2020dynamic,mohamed2020social} and Transformer \cite{aksan2021spatio,wang2021multi} have recently become popular in human motion prediction. These works show significant improvement in capturing the spatial and temporal dependencies of human motion data.

Instead of blindly trusting the predicted human motions, existing studies quantified the uncertainty of the predicted human motions using statistic-based prediction models, e.g., \cite{fridovich2020confidence,park2019planner,faroni2022safety,kanazawa2019adaptive}. These models can naturally predict trajectories in a probabilistic way, handling irregular human movements in HRC. While network-based models typically provide deterministic predictions, some studies have developed techniques to measure the uncertainty inherent in these models and to provide the confidence level associated with the model's outputs. For example, Cheng et al. \cite{cheng2019human} developed a parameter-adaption-based neural network to provide uncertainty bounds of the prediction in real time. Zhang et al. \cite{zhang2020uc} employed conditional variational autoencoders (CVAEs) to sample multiple saliency maps from the latent space, ultimately obtaining an accurate saliency map using the quantified uncertainty. 
Eltouny et al. \cite{eltouny2024tgn} trained an ensemble of motion prediction network models, and estimated the uncertainty based on the aggregation of diverse motion predictions.

\subsection{Robot motion planning}
One of the most important problems in HRC is to plan collision-free robot motions in dynamic workspaces. The computational expense imposed by the curse of dimensionality limits the application of traditional grid-based methods, such as A* algorithm \cite{hart1968formal}. Random-sampling-based methods such as the rapidly exploring random tree (RRT) \cite{lavalle1998rapidly} have demonstrated effectiveness in high-dimensional planning problems. Furthermore, to ensure the optimality of the robot trajectories, asymptotically optimal sampling-based such as batch-informed trees (BIT*) \cite{gammell2015batch}, fast marching trees (FMT*) \cite{janson2015fast}, and optimization-based methods \cite{marcucci2023motion,9001184} are developed. Nowadays, network-based motion planners have been widely used to generate near-optimal robot trajectories with low computational cost. For example, the work in \cite{li2021mpc,bency2019neural,qureshi2020motion} leveraged a network-based model to imitate expert robot trajectories generated from oracle planners, providing near-optimal robot motions. Furthermore, rather than imitating expert trajectories, the work in \cite{cai2021vision,gao2022cola} employed reinforcement learning to obtain the optimal policies to generate robot motions. Despite the advantages demonstrated by such planners, they may struggle to capture the intrinsic connectivity among objects within the workspace. Rather than blindly preprocessing all data together, the graph representation proposed in \cite{liu2024kgplanner} highlights the both local and global dependencies of objects in the workspace when generating robot motions, 
which however does not explicitly consider human motion prediction in the motion planner. 

Incorporating human motion prediction into robotic planning can improve the efficiency of HRC. Cheng et al. \cite{cheng2020towards} included the task recognition and trajectory prediction of human workers into HRC systems to significantly improve efficiency. Unhelkar et al. \cite{unhelkar2018human} proposed a planning algorithm that leverages the prediction of nearby humans to efficiently execute collaborative assembly tasks. Moreover, incorporating the prediction is beneficial for generating collision-free robot trajectories proactively, thus expanding the safety margin of the collaboration. Park et al. \cite{park2019planner} used the predicted human motion to compute collision probabilities for safe motion planning. Kratzer et al. \cite{kratzer2020prediction} proposed a prediction framework that enables the mobile robot to avoid the possible area occupied by a human partner. Zheng et al. \cite{zheng2022human} developed an encoder-decoder network to predict the human hand trajectories, and integrated the avoidance of future collisions as constraints into a model predictive control framework, allowing the planning of safe trajectories.

\section{Uncertainty-Aware Human Motion Prediction}
In this section, we introduce an RNN-based human motion prediction model and explain how the uncertainty of the model is quantified.

\subsection{Human motion predictor}
To predict human trajectories during task execution, we train a prediction model based on an RNN with long short-term memory (LSTM) architecture. Rather than using 3-dimensional position data of arm joints, we employ unit vectors of bones for the network training. In this case, we can ensure a consistent distance between two joints when reconstructing arm poses from bone vectors using the corresponding bone lengths.
This choice preserves the anatomical constraints of the arm during the prediction process. The human arm bone vector is denoted as $x=(\phi_{1},\phi_{2}) \in\mathbb{R}^{6}$, where $\phi_{1}\in\mathbb{R}^{3}$ and $\phi_{2}\in\mathbb{R}^{3}$ are two bone vectors of human upper-arm and forearm respectively. Notably, the position of arm joints and human arm occupied area in the workspace can be reconstructed using $x$ and anthropometric parameters $p_h$, which contain the average bone length and radius of the human arm for each segment. The prediction process is denoted as:
\begin{equation}
    \hat{X}=F(X,\mathbf{W}) \label{LSTM_S}
\end{equation}
where $X{=}[x_{-N+1},...,x_{0}]\in\mathbb{R}^{6N}$ is the human motion of observed $N$ steps, $F(\bullet)$ indicates the prediction function, and $\hat{X}{=}[\hat{x}_{1},...,\hat{x}_m,...,\hat{x}_M]\in\mathbb{R}^{6M}$ stands for the human motion of predicted $M$ steps. Additionally, we treat the well-trained network as the prediction model, and $\mathbf{W}$ indicates the learning weights of the network after training.

\subsection{Uncertainty quantification using MCDS}
The previous subsection briefly introduces that using a network-based motion predictor can predict human trajectories in upcoming time steps. However, human motions in HRC exhibit a certain level of variability that is influenced by factors such as individual characteristics and worker fatigue. Therefore, it's necessary to explicitly quantify uncertainties, enabling effective consideration of variations in human movements. We employ Monte Carlo dropout sampling (MCDS) to quantify the uncertainty of our prediction model, considering that it provides accurate uncertainty estimations and only requires training a single model.

We aim to obtain the prediction distribution such that the possible future trajectories can be utilized for safe robot motion planning. To this end, we apply dropout to every layer of the prediction model, and treat it as a Bayesian approximation of a Gaussian process model over the prediction model parameters \cite{gal2016dropout}. The prediction distribution is calculated using the following equation:
\begin{equation}
p(\hat{X}|X)=\frac{p(\hat{X}|X,\mathbf{W}) p(\mathbf{W})}{p(\mathbf{W}|X,\hat{X})}
\end{equation}
where $p(\mathbf{W})$ is a prior Gaussian distribution over the model parameters, $p(\hat{X}|X,\mathbf{W})$ indicates the likelihood used to capture the prediction process, and $p(\mathbf{W}|X,\hat{X})$ denotes the posterior distribution.

Considering that the posterior distribution can not be evaluated analytically, we use variational inference to approximate it. The approximating distribution $q(\mathbf{W})$ can be close to the true posterior distribution by minimizing the Kullback-Leibler (KL) divergence between them:
\begin{equation}
KL\bigl(q(\mathbf{W})~||~p(\mathbf{W}|X,\hat{X})\bigr)
\end{equation}
where $q(\mathbf{W})$ is defined using Bernoulli distributed random variables and some variational parameters that can be optimized.  
As pointed out in \cite{gal2015bayesian}, the training of the prediction model would also be beneficial for minimizing the KL divergence term. Therefore, $q(\mathbf{W})$ is optimized after the network training, and sampling from $q(\mathbf{W})$ is equivalent to applying dropout on each layer of the prediction model. Eventually, the predictive variance $u$ at test time is calculated using the following equation:

\begin{equation}\label{uncertainty}
\mathbf{u} \approx \frac{1}{K-1} \sum_{k=1}^{K} \left[F(X,\overline{\mathbf{W}}_k)^{T}
F(X,\overline{\mathbf{W}}_k)-K E^{T} E\right]
\end{equation}where $u=[u_1,...,u_m,...,u_M]$ indicates the prediction variance, $K$ is the Monte Carlo sampling size, $\overline{\mathbf{W}}_k$ is fitted to $q(\mathbf{W})$ and denotes the model parameters of the $k$th sample, and $E \approx \frac{1}{K}{\textstyle\sum}_{k=1}^{K} F(X,\overline{\mathbf{W}}_k)$ represents the predictive mean.  

\begin{figure}[h]

	\centering 
	\includegraphics[width=0.75\columnwidth]{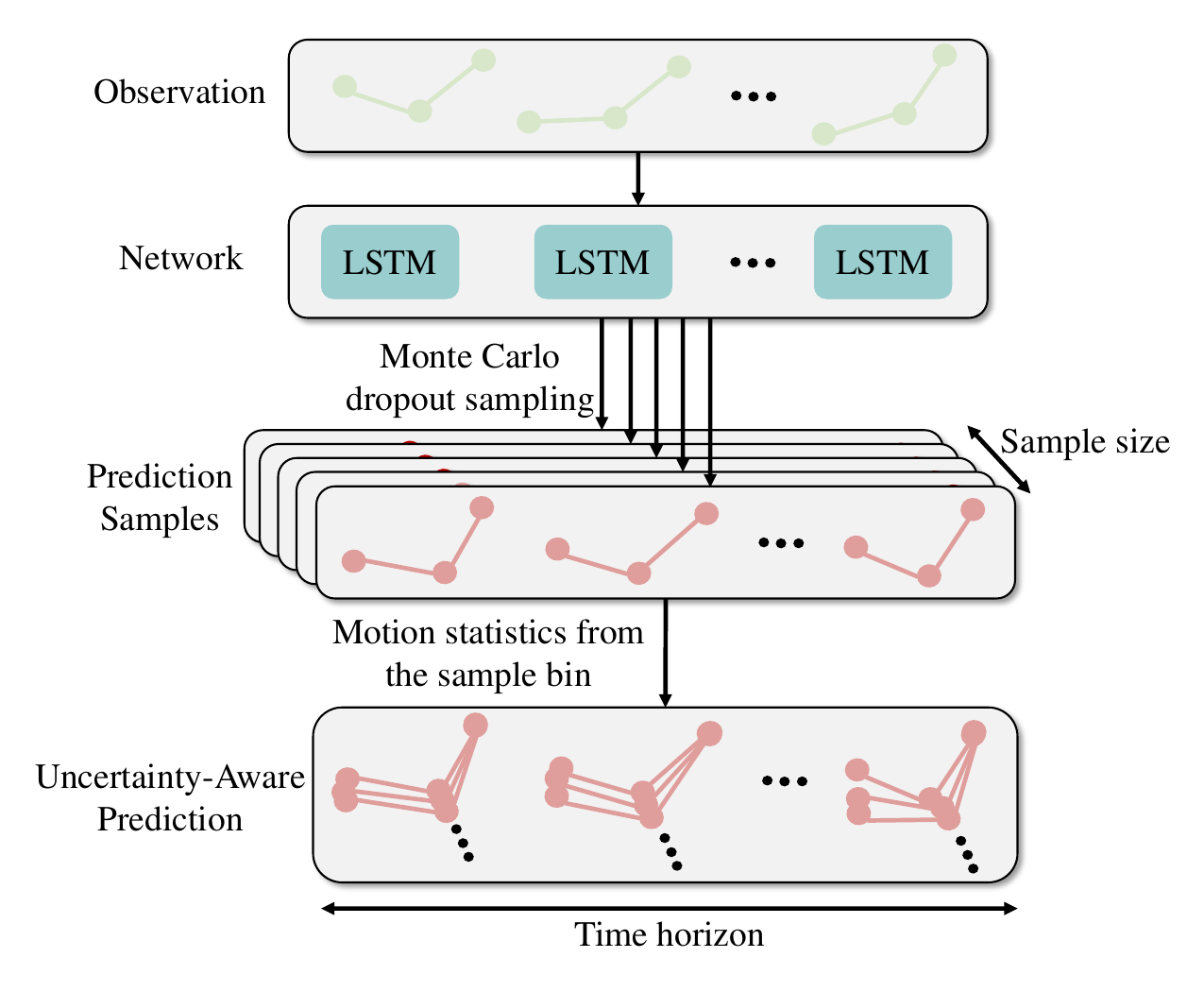}

	\caption{The uncertainty quantification of the prediction model: green dots are the observed human joints, red dots are the predicted human joints, and the uncertainty-aware prediction is generated based on the predictive distribution and includes multiple possible human arm poses at each time step. 
	} \label{fig:quantification}
   \vspace{-0.1in}
\end{figure}

The process of obtaining uncertainty-aware human motion prediction is illustrated in Fig.~\ref{fig:quantification}. The observed human motion is propagated into a well-trained LSTM model. MCDS is employed to generate different possible configurations of the network parameters, and multiple prediction samples are obtained. Finally, the uncertainty-aware prediction includes multiple possible human arm poses at each time step, and is denoted as $\hat{X}^*{=}[\hat{x}_{1}^*,...,\hat{x}_m^*,...,\hat{x}_M^*]\in\mathbb{R}^{6M\times K}$. Notably,  we use $*$ to indicate there are multiple possible arm poses at a predicted time instance. And these poses fit a normal distribution $*{\sim}\mathcal{N} (E,u)$, where $E$ represents the mean and $u$ indicates the predictive variance.

\section{Graph-Based Motion Planner}
This section presents 1) explanations of converting the collaboration workspace and the uncertainty-aware prediction to a graph representation; 2) details of how to leverage a GNN to operate on the constructed graph and generate near-optimal robot motions.

\subsection{Graph representation: illustrating features and connections of objects in the workspace}

\begin{figure}[h]
	\centering 
	\includegraphics[width=0.75\columnwidth]{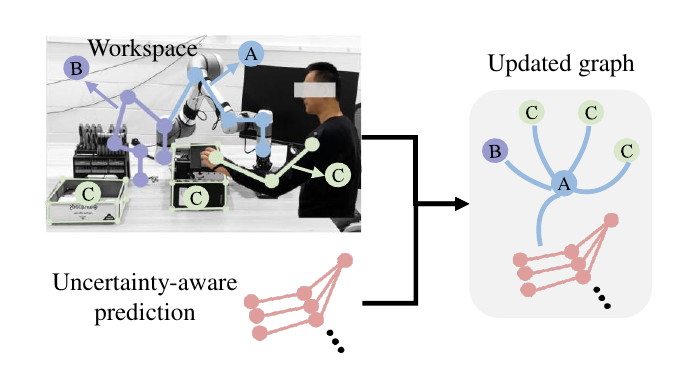}

	\caption{The representation of objects in the workspace and the simplified illustration of the overall graph: the uncertainty-aware prediction is represented using red color. We use dots A, B, and C to simplify the graph representation. Blue dot A indicates the robot's current state, purple dot B denotes the robot's goal state, and green dot C indicates the obstacle's state. All dots contain multiple nodes and edges based on their own structural attributes.
	} \label{fig:Graph}
   \vspace{-0.1in}
\end{figure}

Rather than simply imitating reference motions like traditional neural motion planners, our approach aims to emphasize the dependencies of each key object within the workspace since the object dependencies significantly influence the planning of robot motions.
To highlight such dependencies in the planning, we use nodes to represent the essential objects in the collaboration workspace and connect them using edges. As shown in Fig.~\ref{fig:Graph}, the robot's current state is denoted using six blue nodes, corresponding to the six joints of the robot. The same representation strategy is applied to the robot's goal state and the obstacle states. 
To simplify the illustration, we respectively use dots A, B, and C to represent the robot's current state, the robot's goal state, and the obstacle's state in the graph of Fig.~\ref{fig:Graph}. 
Furthermore, the uncertainty-aware prediction contains multiple future human arm joint positions, which are represented as nodes. In summary, we use $v$ to represent the node of the graph, and $V=[v_1,...,v_t,...,v_T]$ stands for total $T$ essential nodes in the collaboration workspace.

\subsection{Graph operation: node embedding based on neighbors}

In the previous subsection, we employ a graphical representation to efficiently illustrate the objects in the collaborative workspace and showcase their connections. To generate collision-free robot motions, we first employ an oracle planner that generates expert robot trajectories in collaborative workspaces to obtain the training data, and then leverage a GNN to operate on the constructed graph and train the network to generate near-optimal motions.

The graph is described by two matrices: a feature matrix and an adjacency matrix. The feature matrix $H$ describes features of the objects in the workspace, such as the manipulator joint value, the current and future arm's positions, and the static obstacles' potions. The adjacency matrix $A$ indicates the relationships between all nodes. The layers of GNN update features of each node based on the adjacency matrix $A$.  The node embedding process is denoted as: 
\begin{equation}
    h^{(l)}_{v_t}=f_{update}\left( \theta^{(l)},h^{(l-1)}_{v_t},\{h_{v_j}^{(l-1)}\}_{j\in\mathcal{N}_{v_t}}\right)
\end{equation}
where $h^{(l)}_{v_t}$ denotes the embedding of node $v_t$ in the layer $l$, $\theta$ is learning weights, and $\mathcal{N}_{v_t}$ indicates neighbors of node $v_t$. 

\begin{figure}[h]
	\centering 
	\includegraphics[width=0.85\columnwidth]{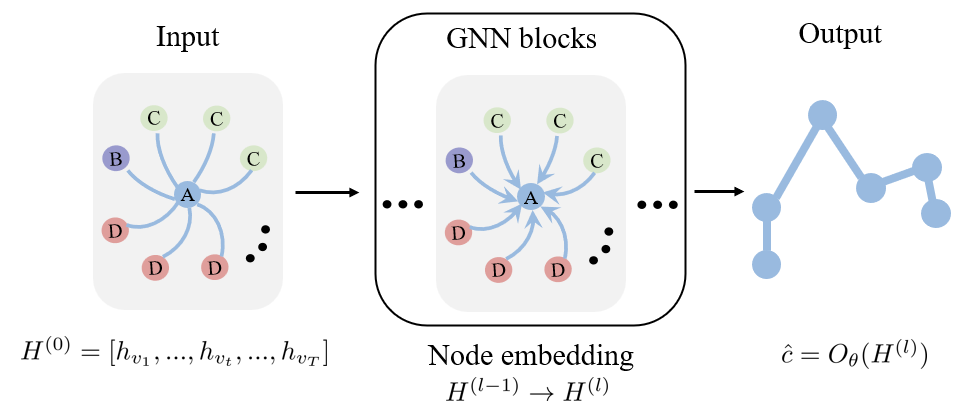}
	\caption{The process of motion generation: red dot D indicates the predicted human arm after MCDS, multiple GNN blocks are used to update the embeddings of nodes, and GNN finally outputs the robot configuration of the next step.
	} \label{fig:Planning}
  \vspace{-0.1in}
\end{figure}

  Fig.~\ref{fig:Planning} illustrates the motion generation using GNN. The red dot D indicates the uncertainty-aware predictions. The GNN input $H^{(0)}=[h_{v_1},...,h_{v_t},...,h_{v_T}]$ is initialized by the node features of the overall graph. All nodes update their embeddings simultaneously in the layer-wise propagation of GNN using the following equation: 
\begin{equation}
H^{(l)}=ReLU\left(\tilde{D}^{-\frac{1}{2}} \tilde{A} \tilde{D}^{-\frac{1}{2}} H^{(l-1)} \Theta^{(l-1)}\right)
\label{eq:update}
\end{equation}
where $ReLU$ is a nonlinear activation function, $H^{(l)}$ indicates the updated feature matrix in the layer $l$, $\tilde{A}=A+I$ is the adjacency matrix with added an identity matrix, $\tilde{D}$ stand for the degree matrix of $\tilde{A}$, and $\Theta^{(l-1)}$ is the learning weights in matrix form.
 Eventually, the output layer takes all updated node embeddings into account and provides the next robot configuration: 
\begin{equation}
\hat{c}=O_{\theta}(H^{(l)})
\end{equation}
where $O_{\theta}$ is the output function, and $\hat{c}$ is the next robot configuration towards the goal region.

\subsection{GNN training}
The previous subsection describes that our neural planner considers various factors for generating the next robot configuration $\hat{c}$ towards the goal position. To ensure the generated motion to be near-optimal, GNN needs to learn optimal paths generated from an oracle planner. The optimal robot path connecting given start and goal configurations is denoted as $\sigma=[c_1,...,c_i,...c_I] \in\mathbb{R}^{I\times d}$, where $d$ is the dimensionality of the robot configuration space.
Utilizing the one-step look ahead planning strategy outlined in \cite{qureshi2020motion}, we define the training loss function for our neural planner as follows:
\begin{equation}
l_{\text {planner}}=\frac{1}{N_z} \sum_z^{N_z} \sum_{i=1}^{I_z-1}\left\|c_{z, i}-\hat{c}_{z, i}\right\|^2
\end{equation}
where $N_z$ is the total number of robot paths in a batch, and $I_z$ is the length of the $z$th path. 

\subsection{Bi-directional planning}

After the network training, the learning weights $\Theta$ in Eq.~(\ref{eq:update}) are well-tuned. We use the well-trained GNN to perform real-time robot motion planning. Based on the one-step-ahead planning strategy, the planned robot configurations are iteratively used as new inputs of our neural planner until a complete path is found.  
Such a planning heavily depends on the previously generated robot configuration, which may potentially accumulate errors throughout the planning process and result in the robot deviating from the goal region. Therefore, we adopt a bi-directional planning way to enhance the robustness of online planning. 

The bi-directional planning starts by initiating two sub-planning branches simultaneously, originating from the start and goal configurations, respectively. Then, we generate linear interpolations trying to directly connect two branches in each planning iteration. Additionally, the two branches grow iteratively until the planned robot configurations and the generated interpolations are both collision-free. Eventually, the two sub-planning branches are stitched together to be a complete robot path. 

  \vspace{-0.1in}

\section{Experimental Validations}

\subsection{Experiment setup}
\subsubsection{Experimental Setting}
\begin{figure}[h]
	\centering 
	\includegraphics[width=0.65\columnwidth]{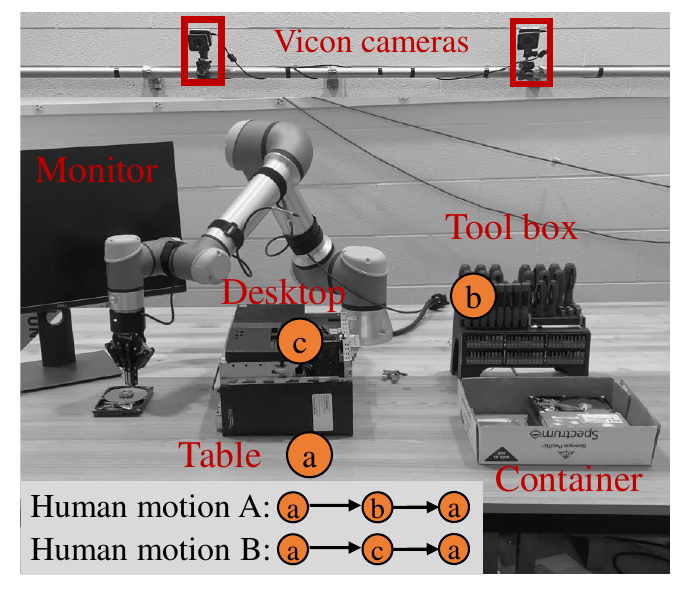}
	\caption{The experimental platform: a, b, and c represent three locations in the table, tool box, and desktop, respectively. Human motion A demonstrates a scenario in which a human worker initially disassembles a hard disk on the table, then reaches towards the tool box to get a new screwdriver, and finally resumes the disassembly task. Human motion B demonstrates a scenario where a human worker initially disassembles a hard disk on the table, then grabs a component from the disassembled desktop, and eventually puts the component on the table.
	} \label{fig:exp_setup}
   \vspace{-0.2in}
\end{figure}

Fig.~\ref{fig:exp_setup} shows the experiment setting of two disassembly scenarios. We use the Vicon motion capture system to track human motions. When constructing the human arm model for collision-checking, we introduce an additional radius to the human arm. This precautionary measure establishes a safety margin between the detected collision and the actual collision.
To ensure safety and prevent potential physical injuries, robot needs to consider both the tracked and predicted human motions.

Note that in the collaboration scenario of this work, the human worker grabs tools located on the workstation while the robot transports disassembled components above the workstation. The physical barrier of the workstation effectively separates the entire human body from the robotic arm. Consequently, the human arm and the robot present the highest likelihood of collisions.
Therefore, tracking and predicting the positions of the forearm and upper-arm suffice to ensure the safety of the collaborative task as outlined in this work. However, it is inadequate for ensuring the safety in scenarios of broader collaboration. Tracking and predicting the movement of the whole human body instead of just the forearm and upper-arm are still necessary for diverse collaborative modes.

\subsubsection{Data acquisition}
We collect 120 human arm trajectories in the frequency of 25Hz for each type of human motion shown in Fig.~\ref{fig:exp_setup}. In our work, one human worker is involved in the data collection, and the human worker is required to perform actions naturally throughout task executions, without deliberate control over the speed. Therefore, while the action speed exhibits some variability, it remains within a reasonable range.
These trajectories are then converted to bone-vectors for network training. We use 70\% of data to train the prediction model. Another 15\% of data is employed for validation, while the remaining portion is reserved for testing. The horizons of the observation and prediction are both 2 seconds. 

\subsubsection{Networks}
The RNN-based human motion prediction model is based on LSTM structure. It consists of three LSTM layers and one dense layer serving as the output layer. Each LSTM layer is followed by a dropout layer with a dropout probability of 10\%. The input dimension of the prediction model is a $50\times6$, where $50$ indicates the observation steps and $6$ implies the number features of arm poses. The output dimension of the model is $50\times5\times6$, where $50$ implies the prediction steps, $5$ indicates the sampling size, and $6$ denotes the number features of the arm pose. 
The GNN consists of five graph convolutional layers employing Rectified Linear Unit activation functions. Subsequently, one global sum pooling layer is applied to compute the sum of node features, and one dense layer is employed as the output layer. The input of the GNN is a constructed graph described with a $55\times6$ feature matrix and a $55\times55$ adjacency matrix, where $55$ indicates the total node number and $6$ implies the number of node features. 

\subsection{Experimental test results}

\subsubsection{Validation of the neural planner}
To evaluate the effectiveness of the graph-based planner, we create a total of 12 different workspaces. We employ RRT* \cite{karaman2011sampling} from the open motion planning library (OMPL) as the oracle planner to generate optimal robot motions in a motion planning platform MoveIt. Each static workspace includes 800 planning scenarios with random pairs of start and goal configurations. 
In static workspaces, we conducted comparative studies between our approach and three other planners from OMPL, which are RRT* \cite{karaman2011sampling}, RRT \cite{lavalle1998rapidly}, and the advanced planner bi-directional FMT* (BFMT* \cite{starek2015asymptotically}). A comparison study is provided in Table~\ref{tab:comparison_detail} \cite{liu2024kgplanner}, from which
the graph planner demonstrates superior performance compared to the other three planners in terms of path length and planning time, and it achieves a promising level of success in generating collision-free motions.

\begin{table}[!]
\footnotesize
	\centering
	\begin{tabular}{cccc}
\toprule[1.5pt]		Planner  & Path length (m) & Planning time (s) & Success rate
\\\toprule[1.5pt]
        RRT  & 1.695 $\pm$ 0.725 & 0.335 $\pm$ 0.650 & 93.5\%\\ 
		
		RRT*($\pm$10\%)   & 1.078 $\pm$ 0.138 & 3.700 $\pm$ 2.421 & 70.2\%\\
  
        BFMT*($\pm$10\%)   & 1.082 $\pm$ 0.137 & 0.541 $\pm$ 0.148 & 91.1\%\\
        
        Graph Planner  & \textbf{1.023 $\pm$ 0.130} &  \textbf{0.197 $\pm$ 0.053} & 88.1\% \\
		\toprule[1.5pt]
	\end{tabular}
	\vspace{5pt}
	\caption{Comparison results of planners: we assume the result data fit normal distribution. Path length refers to the total distance traversed by the manipulator's end-effector along the planned path, planning time quantified the time taken by the planner to compute a collision-free path, and success rate measures the percentage of planning scenarios in which a collision free path is successfully generated from the given start to the goal configuration. Considering the dependency between planning time and trajectory optimality in RRT* and BFMT*, we terminate the planning of RRT* and BFMT* when the planned path length reaches a certain percentage of the path length generated by our approach. Note that the planning time results of RRT is imposed to be normal distribution for better comparison.} 
	\label{tab:comparison_detail}
   \vspace{-0.2in}
\end{table}

\subsubsection{Uncertainty-aware prediction}
We select different values for Monte Carlo sampling size $K$ to obtain the quantified uncertainty. The corresponding results are illustrated in Table.~\ref{tab:sampling_detail}. The ``Elbow/m" and ``Wrist/m" present the standard deviation in predictions relative to the mean predicted joint position. Small values of ``Elbow/m" and ``Wrist/m" indicate greater consistency among multiple arm poses at the predicted time instance, while larger values signify greater variability. Note that there are no target or minimum required values for the quantified uncertainties. A large value signifies increased variability in arm poses at the predicted time instance. This may broaden the scope of possible arm motions for enhanced robot motion planning.
However, due to the close proximity between the potential human arm poses and the robot, it will make the robot more difficult to find motions to avoid potential collisions, and the prolonged inference time increases the risk of human-robot contacts. 
Therefore, the selection of suitable $K$ is a trade-off problem. 
Based on the observation of the Table.~\ref{tab:sampling_detail}, the rise in the value of $K$ leads to a substantial increase in the inference time, whereas the escalation in the quantified uncertainties of elbow and wrist positions has a negligible impact. Therefore, we select $K=5$ to quantify uncertainties and take the quantified uncertainties into the safe robot motion planning. This decision balances the need for comprehensive uncertainty assessment with inference times.

\begin{table}[h]
\scriptsize
	\centering
	\begin{tabular}{cccc}
\toprule[1.5pt]		$K$  & Elbow/m & Wrist/m & Inference time/s  \\\toprule[1.5pt]   5  &  $5.25{\times}10^{-3}$ & $9.95{\times}10^{-3}$ & 0.29\\
        10 &  $5.69{\times}10^{-3}$ ($\uparrow$8.38\%) & $1.01{\times}10^{-2}$ ($\uparrow$1.50\%) & 0.49 ($\uparrow$68.97\%) \\
        20 &  $5.64{\times}10^{-3}$ ($\uparrow$7.43\%)  & $1.01{\times}10^{-2}$ ($\uparrow$1.50\%)  & 1.16 ($\uparrow$300.00\%) \\
		\toprule[1.5pt]
	\end{tabular}

	\caption{Inference time and the standard deviation in predictions based on different sampling sizes, where the value in the parentheses is the increased percentage compared to the selection of $K=5$.}
	\label{tab:sampling_detail}
\end{table}

\subsubsection{Comparison between predictive error and uncertainty}
\begin{figure}[!]
	\centering 	
	\subfigure[The predicted errors of elbow and wrist joint positions]{
	\includegraphics[width=0.8\columnwidth]{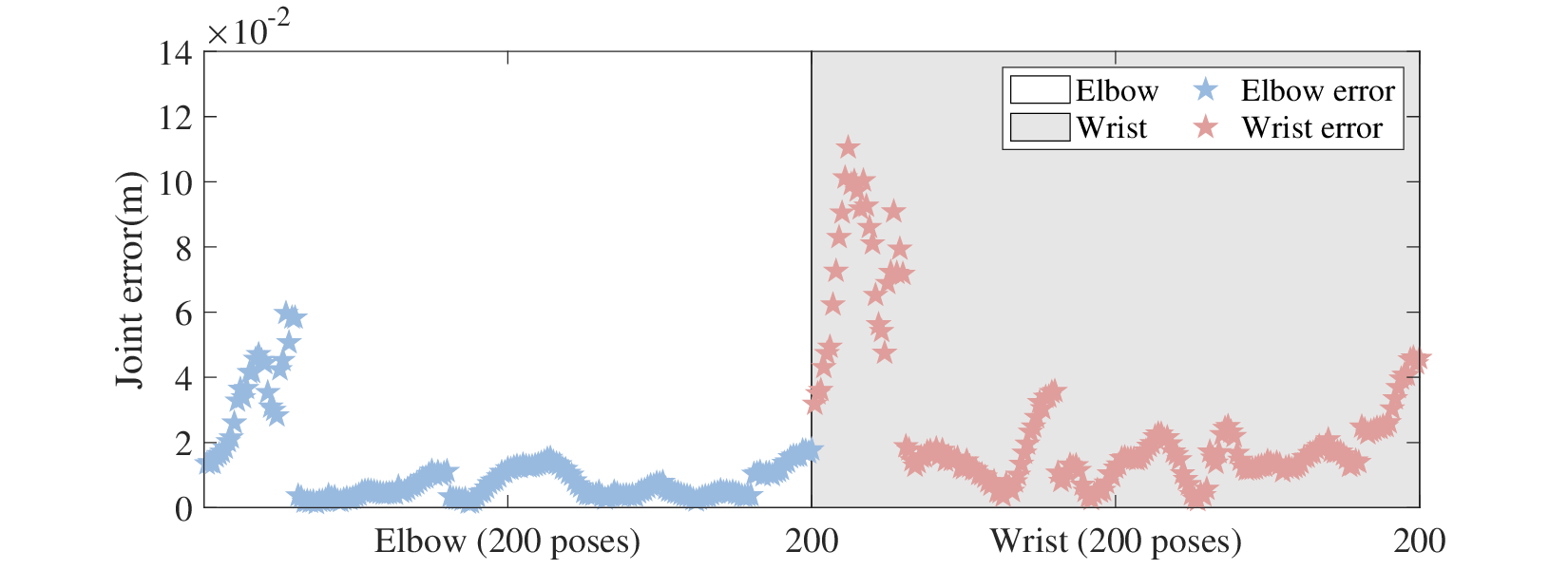}
	\label{fig:errors}
	}
	\subfigure[The quantified uncertainties in terms of elbow and wrist positions]{
	\includegraphics[width=0.8\columnwidth]{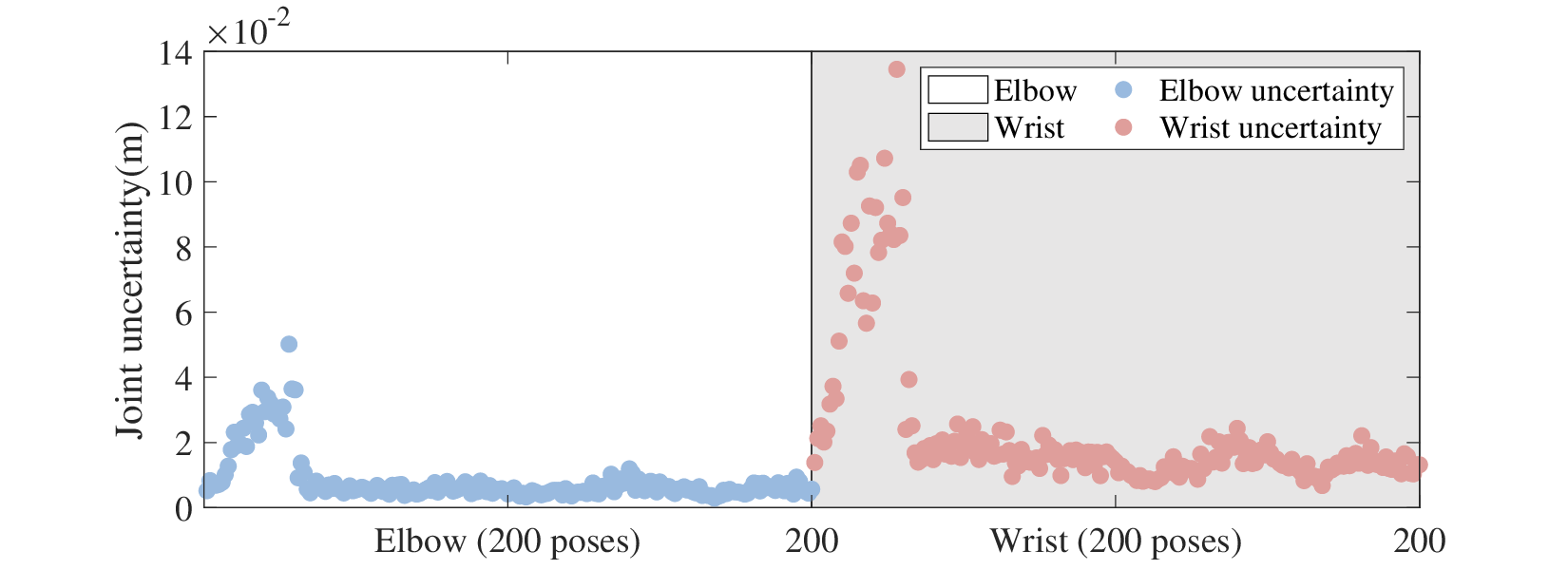}
	\label{fig:uncertainties}
	}
	\caption{The comparison between predicted errors and quantified uncertainties regarding elbow and wrist: high co-relation}
	\label{fig:comparison}
   \vspace{-0.2in}
\end{figure}

We also define the predictive error as the difference between the mean prediction and the ground truth, and compare the quantified uncertainties and predictive errors in terms of the human elbow and wrist joint positions in Fig.~\ref{fig:comparison}. It includes 200 arm poses selected from the test dataset. The predictive error and quantified uncertainty show a high co-relation. Importantly, when human workers are conducting collaborative tasks, the predictive error can not be utilized in the robot planning since it's not feasible to obtain future ground truth at the current time step. Therefore, the quantified uncertainty can be used as an alternative source of information for ensuring human safety.

\begin{figure*}[h]
	\centering 
	\includegraphics[scale=0.38]{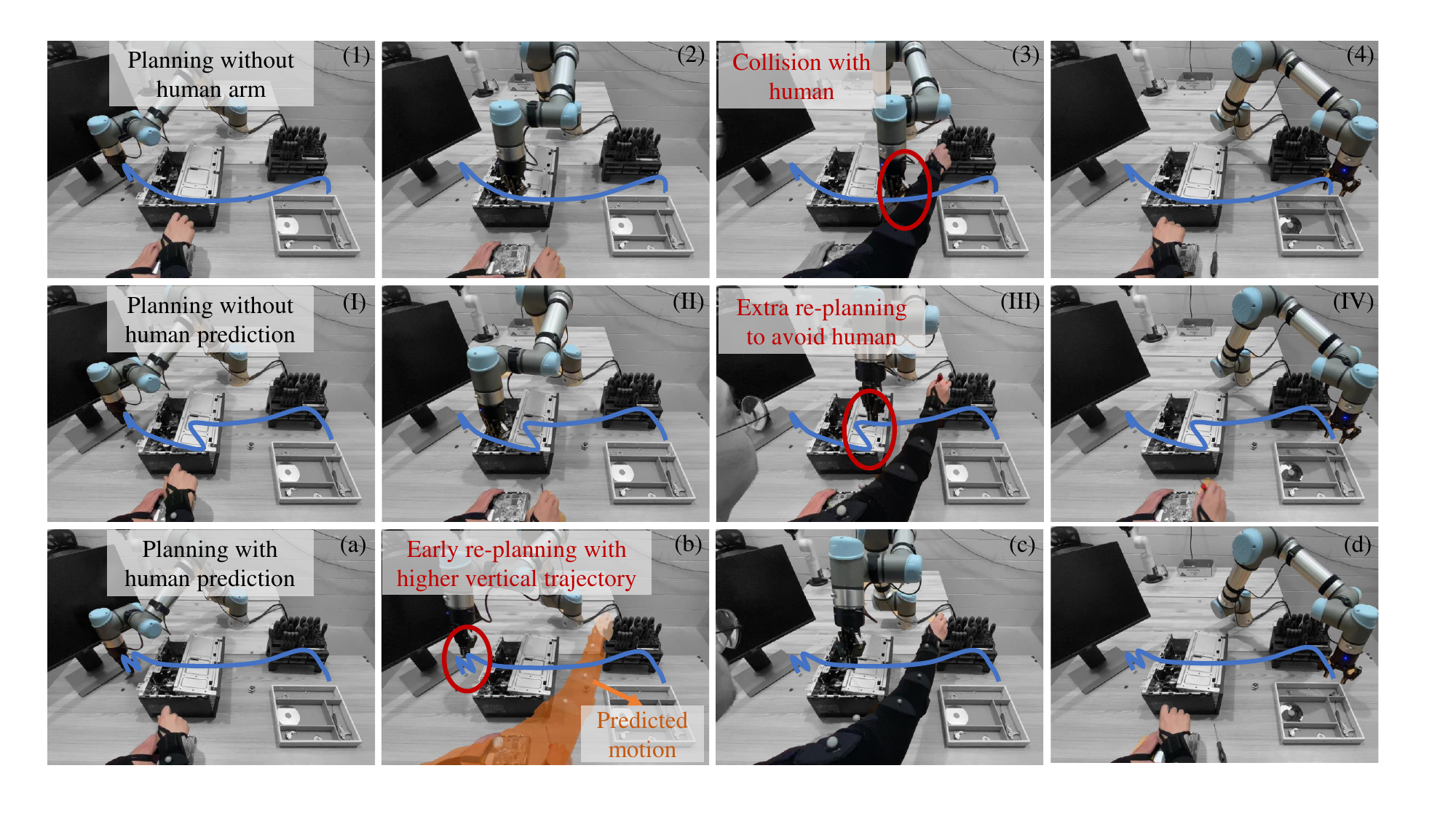}
	\caption{The experimental tests based on human motion A: the experimental tests include three planning cases. Sub-figures (1)$\sim$(4) are the planning scenario without considering the human arm. Sub-figures (I)$\sim$(IV) present the planning based on the current human arm's position, where the robot re-plans its motions to accommodate the reaching motion of the human. Sub-figures (a)$\sim$(d) demonstrate the planning with uncertainty-aware prediction. In sub-figure (b), the observed human motion is putting the screwdriver on the table, and the predicted human motion is reaching for a new screwdriver (i.e., human arm represented by orange color). The robot detects potential future collisions based on the predicted human motion and has an early re-planning to avoid such collisions. Note that the blue end-effector paths are drawn manually.
	} \label{fig:exp1}
   \vspace{-0.2in}
\end{figure*}

\begin{figure*}[!]
	\centering 
	\includegraphics[scale=0.38]{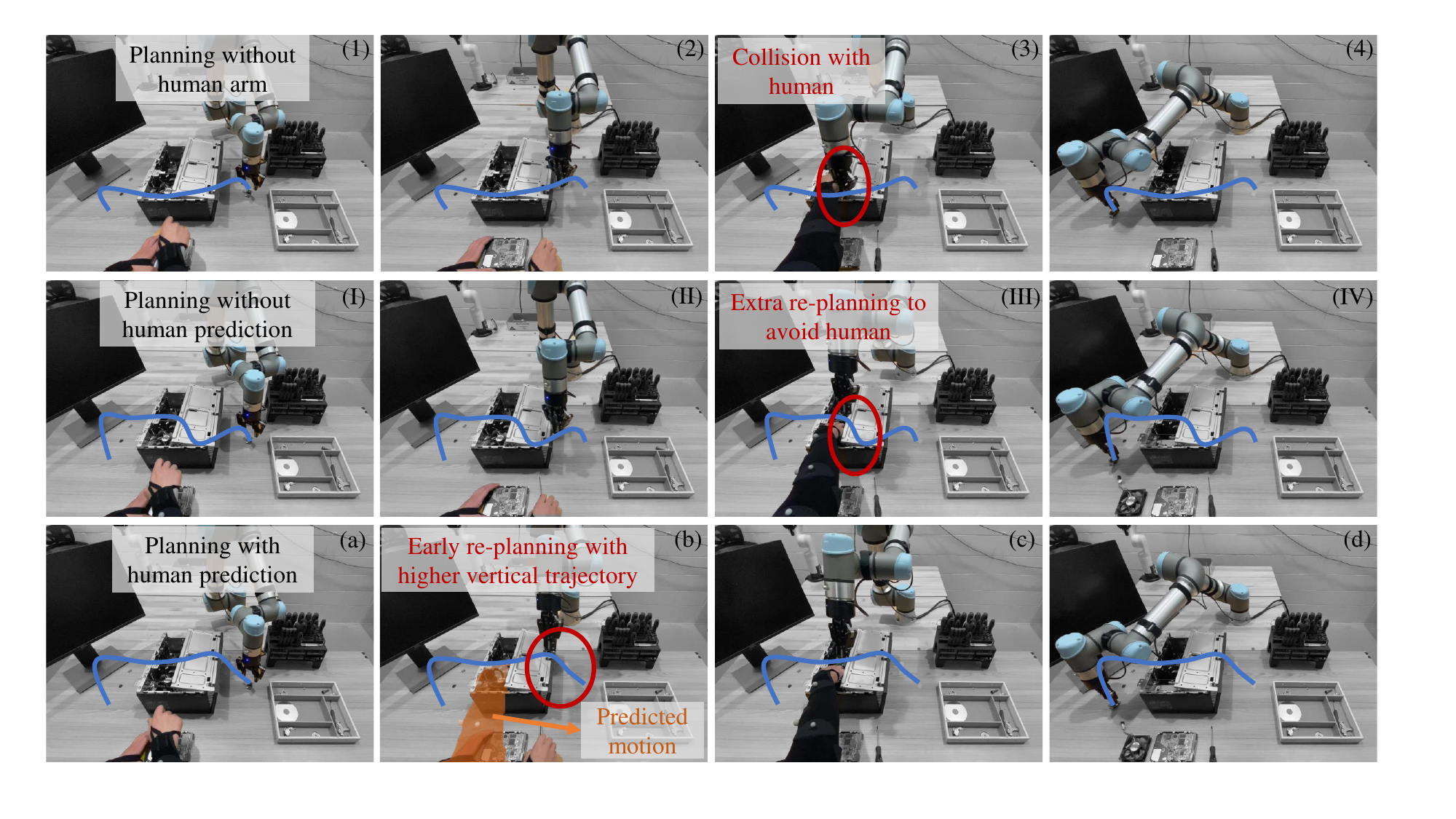}
	\caption{The experimental tests based on human motion B: the experimental tests include three planning cases. Sub-figures (1)$\sim$(4) are the planning scenario without considering the human arm. Sub-figures (I)$\sim$(IV) present the planning based on the current human arm's position, where the robot re-plans its motions to accommodate the grabbing motion of humans. Sub-figures (a)$\sim$(d) demonstrate the planning with uncertainty-aware prediction. In sub-figure (b), the observed human motion is putting the screwdriver on the table, and the predicted human motion is grabbing components from the disassembled desktop (i.e., human arm represented by orange color). The robot detects potential future collisions based on the predicted human motion, and has an early re-planning to avoid such collisions. Note that the blue end-effector paths are drawn manually.
	} \label{fig:exp2}
   \vspace{-0.2in}
\end{figure*}

\subsubsection{Benefits of integrating predictions}
To handle the planning in dynamic workspaces, we simulate a collaborative disassembly scenario and utilize the RRT* planner to generate a set of 12000 collision-free motions for the manipulator in one workspace that involved the current and future human motions. Fig.~\ref{fig:exp1} and Fig.~\ref{fig:exp2} illustrate the experimental tests based on the human motion A and B, respectively. Three cases are considered in the experimental tests: (1) planning without human arm, (2) planning without human prediction, and (3) planning with human prediction. 
The first case is the planning without taking into account the current position of the human arm, resulting in direct contact between the manipulator and the human. Such a case highlights the necessity of the real-time re-planning in HRC scenarios.
In the second case, the planning considers the current position of the human arm.
When the arm is reaching and grabbing components shown in Fig.~\ref{fig:exp1} (III) and Fig.~\ref{fig:exp2} (III), the neural planner is capable of promptly re-planning safe motions to avoid collisions. 
The third case is the planning with uncertainty-aware prediction. Multiple future arm poses are used to check collisions continuously. The manipulator plans motions at the early stage of the task execution since it detects potential collisions according to the predictions. In general, experimental tests show that the robot exhibits abrupt changes in motion when only considering the current human arm positions. On the other hand, by incorporating future human arm poses into the planning process, the robot demonstrates smoother motions in terms of an earlier response and a smoother path for the end-effector.

Quantitative results of smoothness evaluation in terms of velocity profile are provided in Table~\ref{tab:smoothness}. We calculate the acceleration and jerk, and take averages for each step and each robot joint. Without considering the human operator, the robot planner can find a smooth trajectory in the static environment; however, it can lead to collisions since the planner is not human-aware. When considering the human in the environment during the planning process, robot motion will be affected due to the movements of the human operator. However, the smoothness can be improved by incorporating human motion prediction in the planning process, compared to the model without human motion prediction.

\begin{table}[h]
\vspace{-1pt}
\scriptsize
	\centering
	\begin{tabular}{ccc}
\toprule[1.5pt]		Experimental Case  & Acc. ($rad/s^2$) & Jerk. ($rad/s^3$) \\\toprule[1.5pt]   Planning w/o human arm  &  $1.71{\times}10^{-2}$ & $1.94{\times}10^{-2}$\\
        Planning w/o human prediction &  $4.21{\times}10^{-2}$ & $10.1{\times}10^{-1}$\\
        Planning w/ human prediction &  $3.50{\times}10^{-2}$  & $7.58{\times}10^{-2}$ \\
		\toprule[1.5pt]
	\end{tabular}

	\caption{Quantitative results of the smoothness for the robot trajectory. The reported smoothness is evaluated by the average acceleration and jerk (per robot joint per step).}
	\label{tab:smoothness}
   \vspace{-0.2in}
\end{table}

\section{Conclusions and Future Work}
This paper presented a graph-based framework that seamlessly incorporates uncertainty-aware human motion prediction into robotic motion planning. The human motion is predicted using an RNN-based prediction model, and the uncertainty of the prediction model is explicitly quantified using MCDS. The uncertainty-aware prediction is effectively integrated into a graph that represents the collaboration workspace. The manipulator motions are planned based on the constructed graph, and the uncertainty-aware prediction is utilized to expand the safety margin during the planning. The results of the experiments demonstrate that the proposed planning framework can enhance the smoothness and safety of collaborative disassembly processes.  

To further enhance the safety of collaborations, future studies will focus on establishing a target inference time during the collaboration. The inference time can be determined through iterative task execution trials, where the human worker gradually increases the moving speed until contact occurs. 
Additionally, given our heuristic approach of setting an extra radius as a safety distance, establishing the minimum safety distance can also be achieved by targeting this inference time. Additionally, our forthcoming studies will involve reconstructing feasible arm poses at each predicted time instance, empirically collecting uncertainties, and presenting the results in a statistically rigorous manner to better demonstrate the validity of our approach. Furthermore, future studies will validate the smoothness of the robot motions such as the execution velocity and acceleration considering that the path length may not offer a comprehensive measure of optimality.

\ifCLASSOPTIONcaptionsoff
  \newpage
\fi

\bibliographystyle{IEEEtran}
\bibliography{ref}{}

\end{document}